\documentclass[pdflatex,sn-mathphys-num]{sn-jnl}

\usepackage{graphicx}%
\usepackage{multirow}%
\usepackage{amsmath,amssymb,amsfonts}%
\usepackage{amsthm}%
\usepackage{mathrsfs}%
\usepackage[title]{appendix}%
\usepackage{xcolor}%
\usepackage{textcomp}%
\usepackage{manyfoot}%
\usepackage{booktabs}%
\usepackage{algorithm}%
\usepackage{algorithmicx}%
\usepackage{algpseudocode}%
\usepackage{listings}%
\usepackage{graphics} 
\usepackage{graphicx}
\usepackage{epsfig} 
\usepackage{mathptmx} 
\usepackage{times} 
\usepackage{amssymb}  
\usepackage{microtype}
\usepackage{algorithm}
\usepackage{float}
\usepackage{makecell}

\theoremstyle{thmstyleone}%
\theoremstyle{thmstyletwo}%

\theoremstyle{thmstylethree}%

\begin{document}

\title{Topology Optimization of Leg Structures for Construction Robots Based on Variable Density Method}


\author[1]{\fnm{Xiao} \sur{Liu}}\email{xiao.liu1@siat.ac.cn}

\author[2]{\fnm{Xianlong} \sur{Yang}}\email{yang.xl@siat.ac.cn}

\author[1,3]{\fnm{Weijun} \sur{Wang}}\email{wj.wang@giat.ac.cn}

\author*[1,3]{\fnm{Wei} \sur{Feng}}\email{wei.feng@siat.ac.cn}

\affil[1]{\orgdiv{Shenzhen Institute of Advanced Technology}, \orgname{Chinese Academy of Sciences}, \orgaddress{\street{No. 1068 Xueyuan Avenue}, \city{Shenzhen}, \postcode{518055}, \state{Guangdong}, \country{China}}}

\affil[2]{\orgdiv{School of Transportation and Logistics Engineering}, \orgname{Wuhan University of Technology}, \orgaddress{\street{No. 122, Heping Avenue}, \city{Wuhan}, \postcode{430070}, \state{Hubei}, \country{China}}}

\affil[3]{\orgname{University of Chinese Academy of Sciences}, \orgaddress{\street{No. 1 Yanqi Lake East Road}, \city{Beijing}, \postcode{101408}, \country{China}}}


\abstract{In complex terrain construction environments, there are high demands for robots to achieve both high payload capacity and mobility flexibility. As the key load-bearing component, the optimization of robotic leg structures is of particular importance. Therefore, this study focuses on the optimization of leg structures for construction robots, proposing a topology optimization strategy based on the SIMP (Solid Isotropic Microstructures with Penalization) variable density method along with a structural re-design approach. The design performance is comprehensively validated through finite element analysis using ANSYS. First, static and modal analyses are conducted to evaluate the rationality of the initial design. Then, topology optimization using the SIMP-based variable density method is applied to the femur section, which accounts for the largest proportion of the leg's weight. Based on iterative calculations, the femur undergoes secondary structural reconstruction. After optimization, the mass of the femur is reduced by 19.45\%, and the overall leg mass decreases by 7.92\%, achieving the goal of lightweight design. Finally, static and modal analyses are conducted on the reconstructed leg. The results demonstrate that the optimized leg still meets structural performance requirements, validating the feasibility of lightweight design. This research provides robust theoretical and technical support for lightweight construction robot design and lays a foundation for their efficient operation in complex construction environments.}




\maketitle

\section{Introduction}

In modern construction environments, the application of robotics is increasingly widespread, particularly in heavy-load and complex construction tasks, where construction robots demonstrate significant advantages [1,2,3]. These tasks impose higher demands on the payload-to-weight ratio of construction robots. Excessive self-weight not only reduces mobility and flexibility but also increases energy consumption, impacting operational efficiency and applicability. Therefore, achieving lightweight design while ensuring structural strength and performance has become a critical research topic in the field of construction robotics.

Current construction robots predominantly adopt wheeled [4,5], tracked [6,7], or rail-based mobility systems [8,9], which, while stable on flat terrain, are limited by weak terrain adaptability, insufficient flexibility, and high spatial constraints. With the continuous integration of technology and information systems, the new generation of construction robots achieves significant reductions in construction costs and risks through automation and intelligent design, while efficiently operating in hazardous environments such as narrow or dark spaces, thereby enhancing construction efficiency and safety. To adapt to construction sites with numerous obstacles and complex terrains, we propose a hexapod mobile chassis for construction robots, as shown in Figure 1. Consequently, as a critical structural component, the robot legs require structural analysis and lightweight design to achieve a lower self-weight and enhanced mobility.

Current research on robot lightweight design primarily focuses on materials and structural optimization. Among these, topology optimization stands out as a critical approach for achieving lightweight structures due to its greater design freedom and broader design space [10,11,12,13]. In recent years, researchers have proposed various topology optimization methods for continuum structures, including the homogenization method [14], variable thickness method [15], evolutionary structural optimization (ESO) method [16], independent continuous mapping (ICM) method [17,18], level set method [19], phase field method [20], moving morphable components (MMC) method [21], and topological derivative method [18,22]. Each of these methods has unique characteristics and demonstrates strong applicability and effectiveness in addressing different types of structural optimization problems.  Despite significant advancements in accuracy, computational efficiency, and shape representation capabilities, each method has its limitations. For instance, the homogenization method is suitable for theoretical research but faces challenges in practical manufacturing integration. The level set method excels in interface control but struggles with complex geometries and boundary conditions. The phase field method shows advantages in handling topological changes but incurs relatively high computational costs. In contrast, the variable density method, known for its efficiency, rapid computation, and high adaptability to geometric constraints in design domains, emerges as an ideal choice for constructing high-performance structural layouts. Therefore, for the structural optimization of construction robot legs, we adopt a strategy based on the SIMP variable density method.

Although topology optimization methods have achieved significant success in fields such as aerospace, automotive, and industrial robotics [23,24,25,26,27], their application in structural optimization for specific operating conditions and complex environments remains challenging. In the design of construction robot legs, in particular, the structure must not only exhibit excellent lightweight characteristics but also meet strength, stiffness, and dynamic performance standards under various operating loads. However, current research often lacks sufficient consideration of feasibility validation, performance evaluation of optimized results, and multi-scale reconstruction of structures. Existing methods may fail to address all requirements, thereby limiting the engineering implementation and broader application of optimization solutions.
\begin{figure}[h]
\centering
\includegraphics[width=0.6\textwidth]{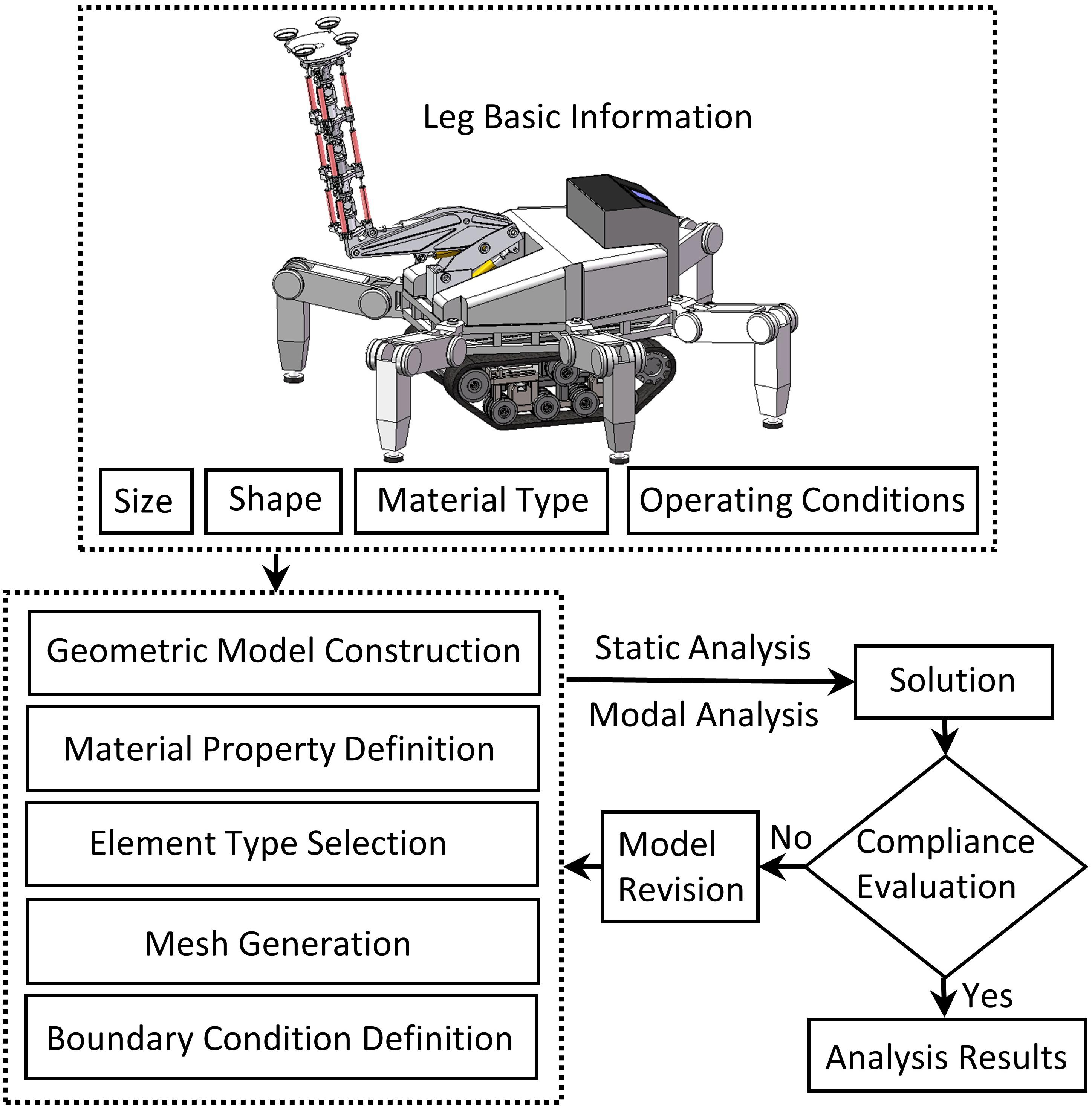}
\caption{Flowchart of Leg Finite Element Analysis}\label{Figure 1}
\end{figure}

To address the unique demands of structural optimization for construction robot legs, this study proposes a topology optimization and secondary structural reconstruction strategy based on the SIMP variable density method, aiming to explore its performance and applicability in construction scenarios. This approach seeks to expand the application scope of topology optimization methods while providing new perspectives and technical directions for lightweight design, thereby promoting advancements in the construction field. Compared to traditional wheeled or tracked construction robots, the robot studied here incorporates a multi-legged design to enhance adaptability in complex terrains with numerous obstacles. As a critical structural component, the legs exhibit significant research value and optimization potential. To meet the operational requirements of mobile construction robots while ensuring load-bearing safety during tasks, reducing the overall weight becomes a key consideration in the design process. To address this challenge, this study focuses on the leg structure, employing the SIMP-based variable density method for topology optimization and conducting detailed static and modal analyses using ANSYS. The aim is to achieve structural lightweighting while maintaining performance. This research not only provides new insights into improving the mobility and flexibility of construction robots but also lays a solid foundation for their operation in more challenging environments.
\section{Finite Element Analysis of Construction Robot Legs}
\subsection{Finite Element Analysis of the Leg}
The Finite Element Method (FEM) is a widely used numerical technique for solving multi-physics problems. By discretizing a complex domain into a finite number of elements connected via nodes, approximate solutions are obtained using interpolation functions within each element. These elements are then reassembled into a global system through boundary conditions to derive an approximate solution for the complex domain. Finite Element Analysis (FEA) enables static analysis, modal analysis, and topology optimization of complex structures. The FEA process typically consists of three main steps: pre-processing, solving, and post-processing. The flowchart of the FEA process for the leg in this study is shown in Figure 1.

The specific analysis steps are as follows:  
(1) Determine the basic parameters of the leg, including its shape, dimensions, working conditions, and material type.  
(2) Simplify the leg structure, create a geometric model, and define the material properties of the structure.  
(3) Based on the geometric characteristics of the leg, select the type and number of mesh elements, generate the mesh for the leg model, and verify the mesh quality.  
(4) Define boundary conditions, including loads and displacement constraints.  
(5) Perform solution tasks, such as static analysis and modal analysis.  
(6) Post-process the computational results and refine the model or adjust pre-processing settings iteratively until reasonable and accurate results are obtained.

This construction robot employs a tripod gait during locomotion, with three legs in the stance phase and the other three in the swing phase. During movement, the legs in the stance phase primarily support the entire weight of the robot and bear the maximum loads from various directions, compared to the legs in the swing phase. Therefore, finite element analysis is conducted on the legs in the stance phase, and topology optimization of the legs is performed based on the analysis results.
\subsection{Static Analysis of the Leg in the Support Phase}

The leg model is imported into ANSYS Workbench, with the main body of the leg made of 6061 aluminum alloy. The material has a density of 2.77 g/cm³, a Poisson's ratio of 0.330, and an elastic modulus of 69.6 GPa. According to the technical specifications of the robot, the total load capacity of the hexapod construction robot is 400 kg, and the maximum torque at the root joint is 300 N·m. Assuming the robot walks in a tripod gait, the maximum load on a single leg on one side is 200 kg, which represents the worst-case condition for the supporting leg.   Given a safety margin factor of 1.2, a vertical downward force of 2400 N and a torque of 360 N•m around the joint axis are applied at the distal root joint, with the global standard gravitational acceleration set to 9.8066 m/s². Assuming that the leg is in the support phase during walking, the foot end is stationary relative to the ground, and thus the degrees of freedom of the foot end are constrained.

After completing the material parameter settings, mesh generation, and boundary condition establishment in ANSYS Workbench, the analysis task is created and submitted for computation. The static analysis results of the supporting leg are obtained. The displacement contour and Von-Mises equivalent stress contour of the leg are shown in Figure 2.

\begin{figure}[h]
\centering
\includegraphics[width=0.6\textwidth]{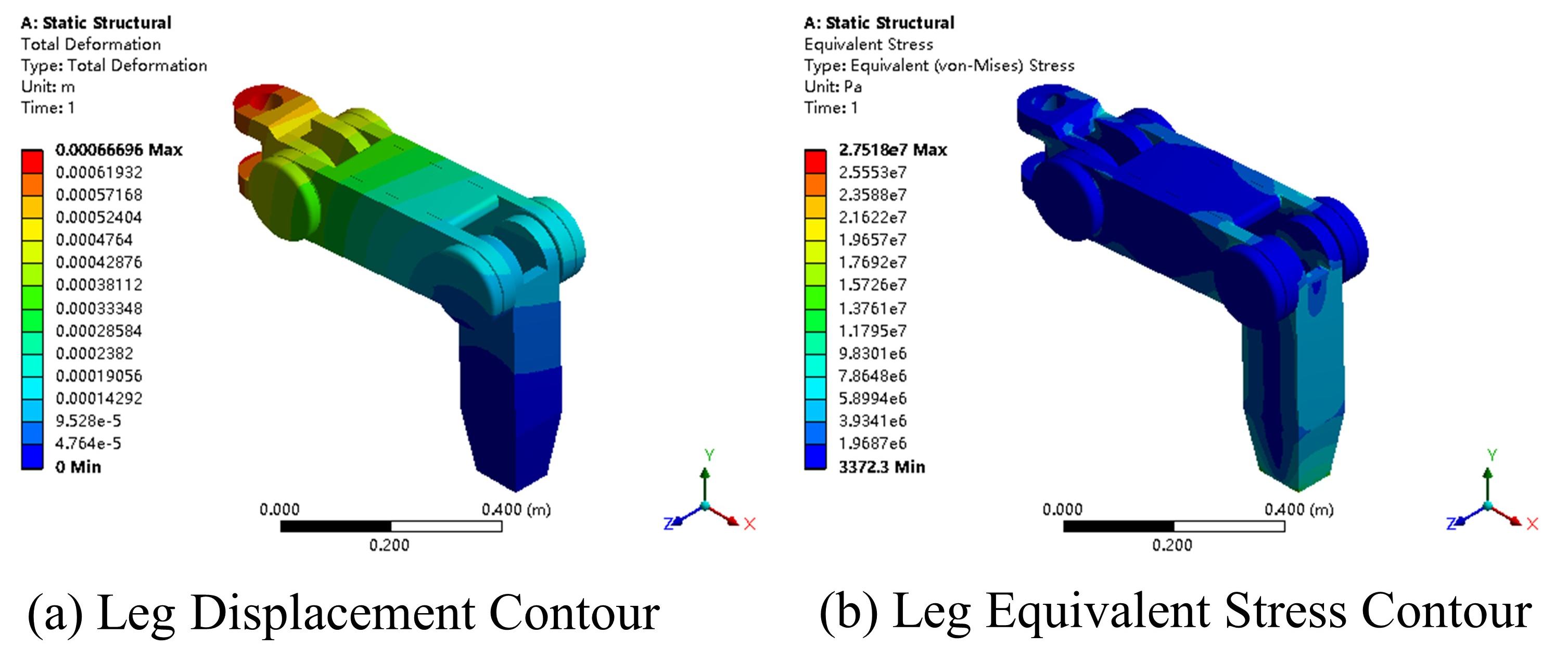}
\caption{Results of Static Analysis of the Leg}\label{Figure 2}
\end{figure}

The static analysis results of the femur segment are shown in Figure 3.

\begin{figure}[h]
\centering
\includegraphics[width=0.6\textwidth]{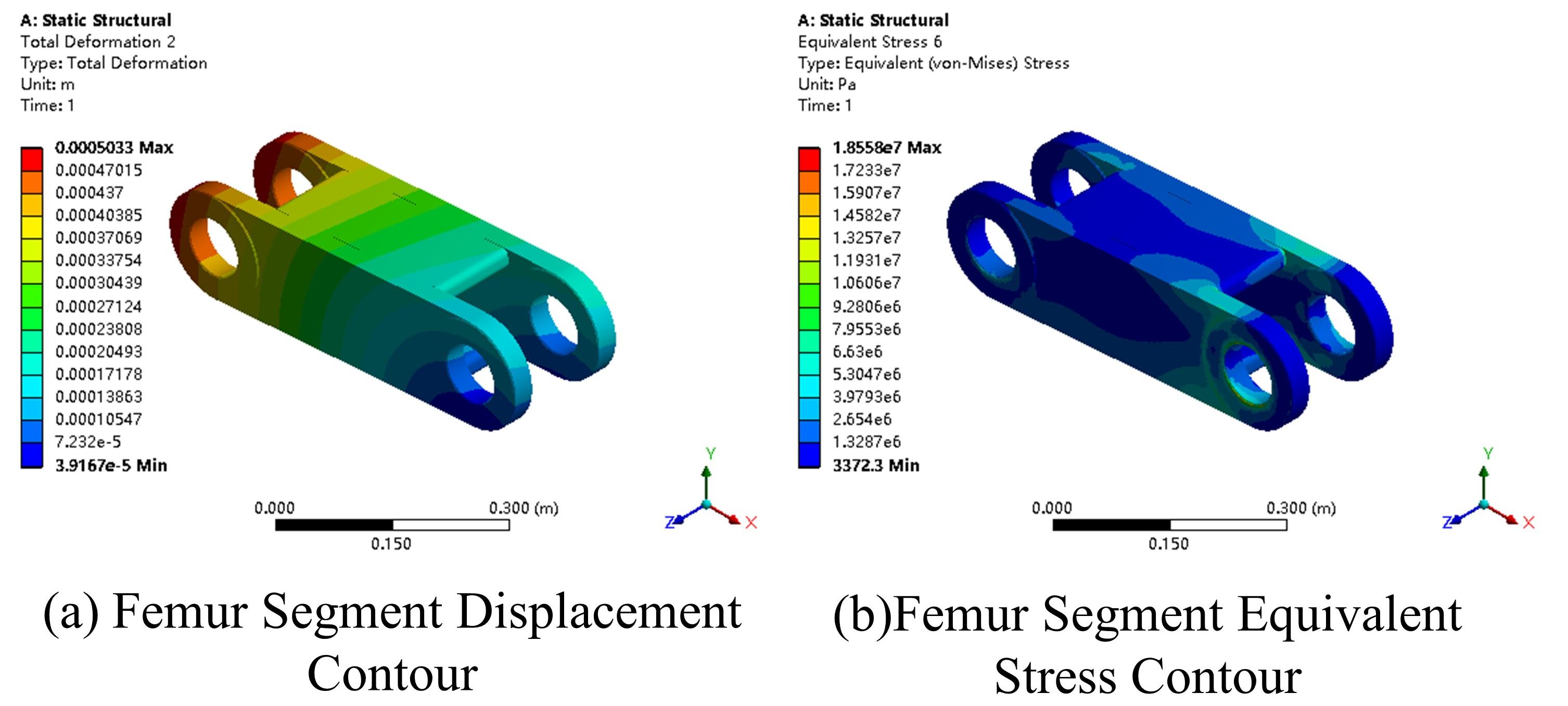}
\caption{Results of Static Analysis of the Femur Segment}\label{Figure 3}
\end{figure}
The analysis results show that, when the leg is in the supporting phase, the maximum equivalent stress of the entire leg is 27.52 MPa, and the maximum equivalent stress on the femur segment is 18.56 MPa. Stress singularities primarily occur at the joints of each leg segment. The yield strength of 6061 aluminum alloy is 276 MPa, and with a safety factor of 2, the allowable stress is 138 MPa. The maximum equivalent stress of both the overall leg and the femur segment is below the allowable stress, thus satisfying the strength requirements. The maximum displacement of the entire leg is 0.67 mm, and the maximum displacement of the femur segment is 0.50 mm, both of which meet the precision requirements. These results preliminarily validate the rationality of the leg's structural design.

\subsection{Modal Analysis of the Leg}\label{subsubsec2}
During operation, the structural components of the construction robot's leg vibrate. To ensure the safety of the leg during the supporting phase, a modal analysis is performed to determine the structural vibration characteristics of the leg. Modal analysis of the leg in the support phase is conducted using ANSYS Workbench to determine the natural frequencies and mode shapes of the first six modes. The first six mode shapes are illustrated in Figure 4.

\begin{figure}[h]
\centering
\includegraphics[width=0.6\textwidth]{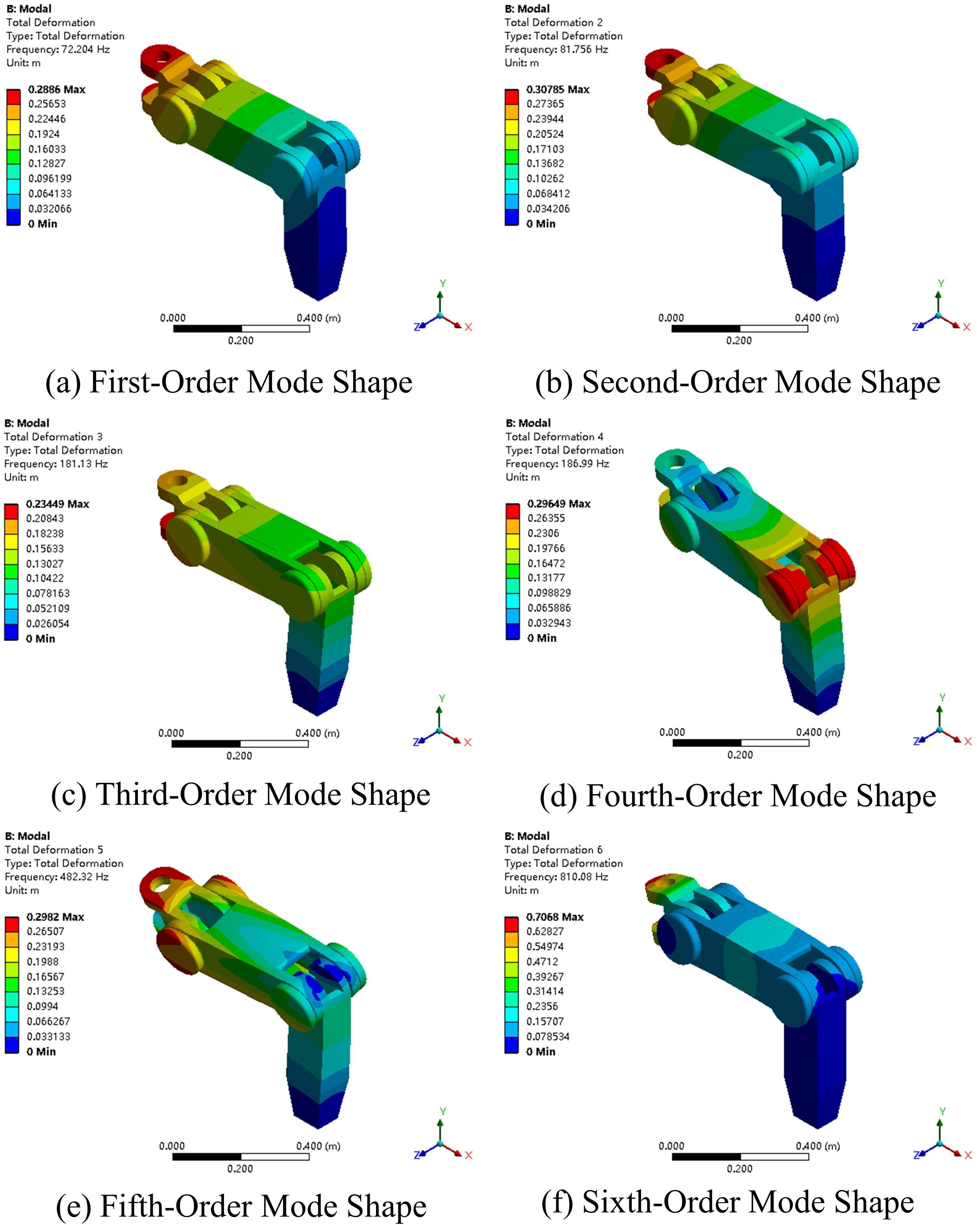}
\caption{Contour Plots of the First Six Modes of the Leg}\label{Figure 4}
\end{figure}
The natural frequencies and mode shapes of the first six modal orders of the leg, as determined from the analysis results, are summarized in Table 1.

\begin{table}[ht]
\centering
\caption{The First Six Mode Shapes of the Leg}
\label{tab:mode_shapes}
\begin{tabular}{ccc}
\toprule
Order & Natural Frequency (Hz) & Mode Shape \\
\midrule
1 & 72.204 & Bending in XOZ Plane \\
2 & 81.756 & Bending in XOY Plane \\
3 & 181.13 & Bending in YOZ Plane \\
4 & 186.99 & Combined Bending and Torsion \\
5 & 482.32 & Combined Bending and Torsion \\
6 & 810.08 & Combined Bending and Torsion \\
\bottomrule
\end{tabular}
\end{table}

As shown in Table 1, the natural frequency of the first-order mode is 72.204 Hz, and the natural frequencies increase with the modal order, requiring a higher number of nodes to induce resonance. Therefore, for the unoptimized leg, although the mode shapes differ, the natural frequencies of all modal orders are higher than the operational frequency of the leg. Consequently, resonance is unlikely to occur during the leg's operation.

\section{Topology Optimization Design of the Leg Femur Section}

As indicated by the static and modal analyses, the femur segment of the leg exhibits significant potential for optimization and accounts for the highest proportion of the total weight. Therefore, topology optimization is performed on the femur segment of the leg.
\subsection{Topology Optimization Methods and Procedures for Continuous Structures}\label{subsubsec2}
In topology optimization, there are three essential elements: the objective function, design variables, and constraints. The general mathematical model for these elements is expressed as follows:

Objective function:
\begin{equation}
\label{deqn_ex1}
\mathrm{min~f(x)}
\end{equation}
In the equation, \( f(x) \) represents the objective function for topology optimization, which is typically related to the structure's compliance, mass, or dimensions.

Design variables:
\begin{equation}
\label{deqn_ex1}
X=
\begin{bmatrix}
X_1X_2...X_n
\end{bmatrix}^T
\end{equation}
In the equation, \( X \) represents the design variables for topology optimization, which can be a single or multiple design variables.

Constraints:
\begin{equation}
\label{deqn_ex1}
\begin{cases}
g_i\left(x\right)\leq0,i=1,2,...,k \\
h_j\left(x\right)=0,j=1,2,...,l & 
\end{cases}
\end{equation}

In the equation, \( g_i(x) \) and \( h_j(x) \) represent the inequality and equality constraints, respectively, in the optimization process. The overall workflow for topology optimization design is shown in Figure 5.

\begin{figure}[h]
\centering
\includegraphics[width=0.6\textwidth]{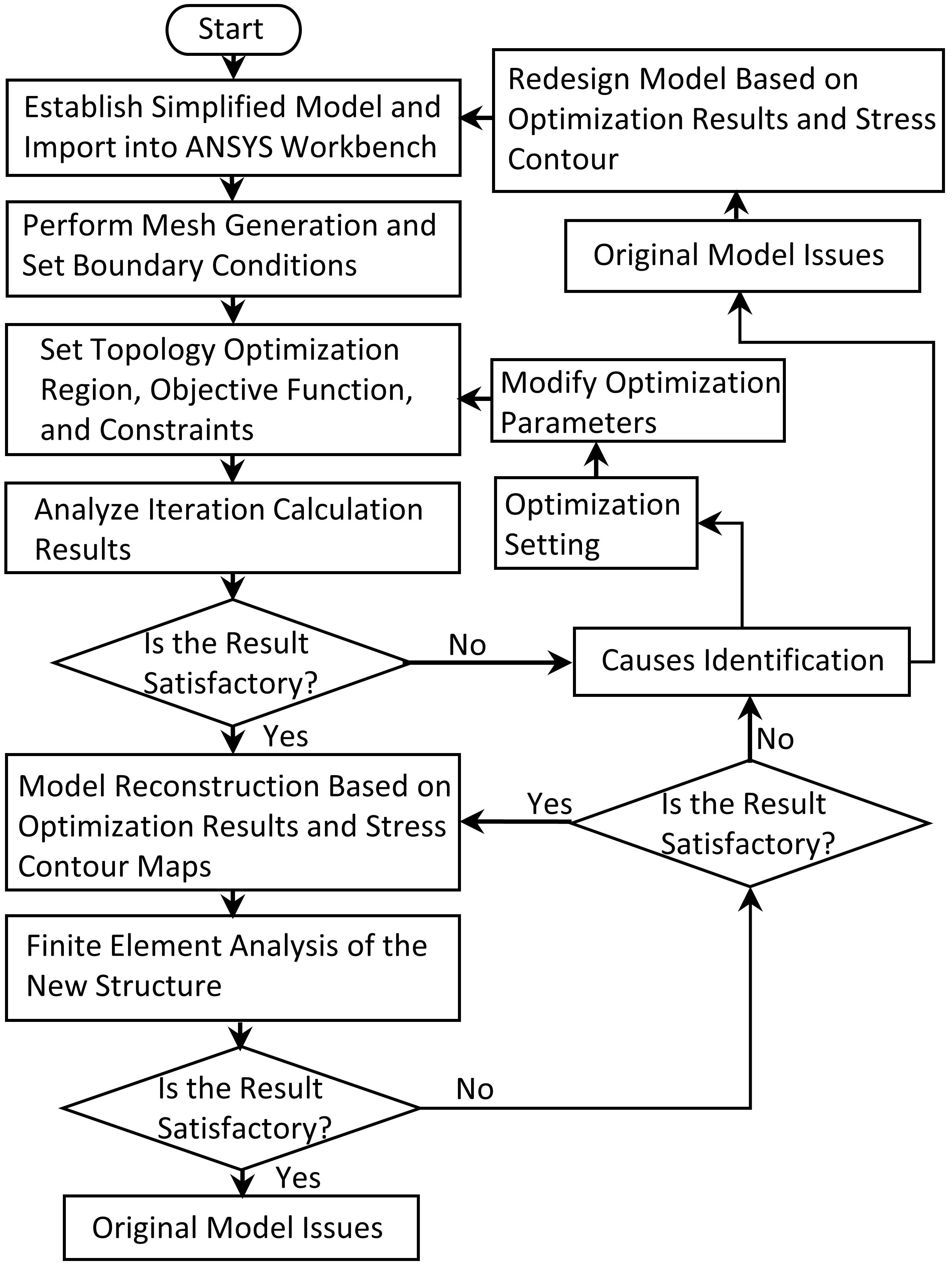}
\caption{Overall Flowchart of the Topology Optimization Design}\label{Figure 5}
\end{figure}
\subsection{Topology Optimization Based on the SIMP Variable Density Method}\label{subsec2}

The variable density method optimizes the topology by considering the density of each element in the model, transforming the structural topology optimization problem into an optimal material distribution problem [28]. Typically, the Solid Isotropic Material with Penalization (SIMP) model [29] is employed, which introduces a penalization interpolation factor to suppress intermediate densities in the design elements. This confines the density values of the elements to either 0 or 1, rather than continuous variation. Elements with a density value of 0 are removed, while those with a density value of 1 are retained, thereby achieving material elimination. The mathematical model of the SIMP variable density method is expressed as follows:

\begin{equation}
\label{deqn_ex1}
\begin{cases}
Min:C\left(X\right)=U^TKU=\sum_{e=1}^N\left(x^e\right)^pu_e^Tk_0u_e \\
Subjectto:\frac{V\left(X\right)}{V_0}\leq f \\
KU=F \\
0<x_{\min}\leq x^e\leq x_{\max}\leq1 & 
\end{cases}
\end{equation}

In the equation, \( C(X) \) represents the compliance of the structure, which is the objective for optimization design; \( U \) and \( K \) are the overall displacement matrix and stiffness matrix, respectively, with \( F \) being the load matrix; \( x_e \) is the design variable for an element, representing the density of the element, where \( P \) is the penalization factor with \( P > 0 \); \( u_e^T \) and \( k_0 \) are the element displacement matrix and stiffness matrix, respectively; \( V(X) \) and \( V_0 \) are the effective volumes of the structure before and after optimization, respectively, with \( f \) being the expected volume optimization coefficient.
\subsection{Topology Optimization Results Analysis}

The femur segment of the leg adopts a box-type structure, primarily composed of side plates perpendicular to the horizontal plane and vertical plates parallel to the horizontal plane. During the walking process of the construction robot, the vertical plates bear not only the robot's mass load but also the varying loads required to drive its forward motion, resulting in complex stress conditions. To ensure the reliability of the supporting leg during the robot's walking, only the side plates of the femur segment are selected as the optimization targets in the topology optimization process. Additionally, the regions where the side plates contact other structural components are designated as non-optimizable areas.

Based on the static analysis results of the femur segment, the optimization target for the design domain is set to achieve a 50\% reduction in material volume. Following the optimization design process outlined in Figure 5 and the static analysis results of the supporting leg, the Topology Optimization module in ANSYS Workbench is utilized to perform topology optimization on the femur segment using the SIMP-based variable density method. After multiple modifications to the model and adjustments to the optimization parameters, the final result is obtained after 44 iterations, as shown in Figure 6. In the figure, the gray regions represent retained material, the red regions indicate material to be removed, and the yellow regions denote transitional boundaries.

\begin{figure}[h]
\centering
\includegraphics[width=0.6\textwidth]{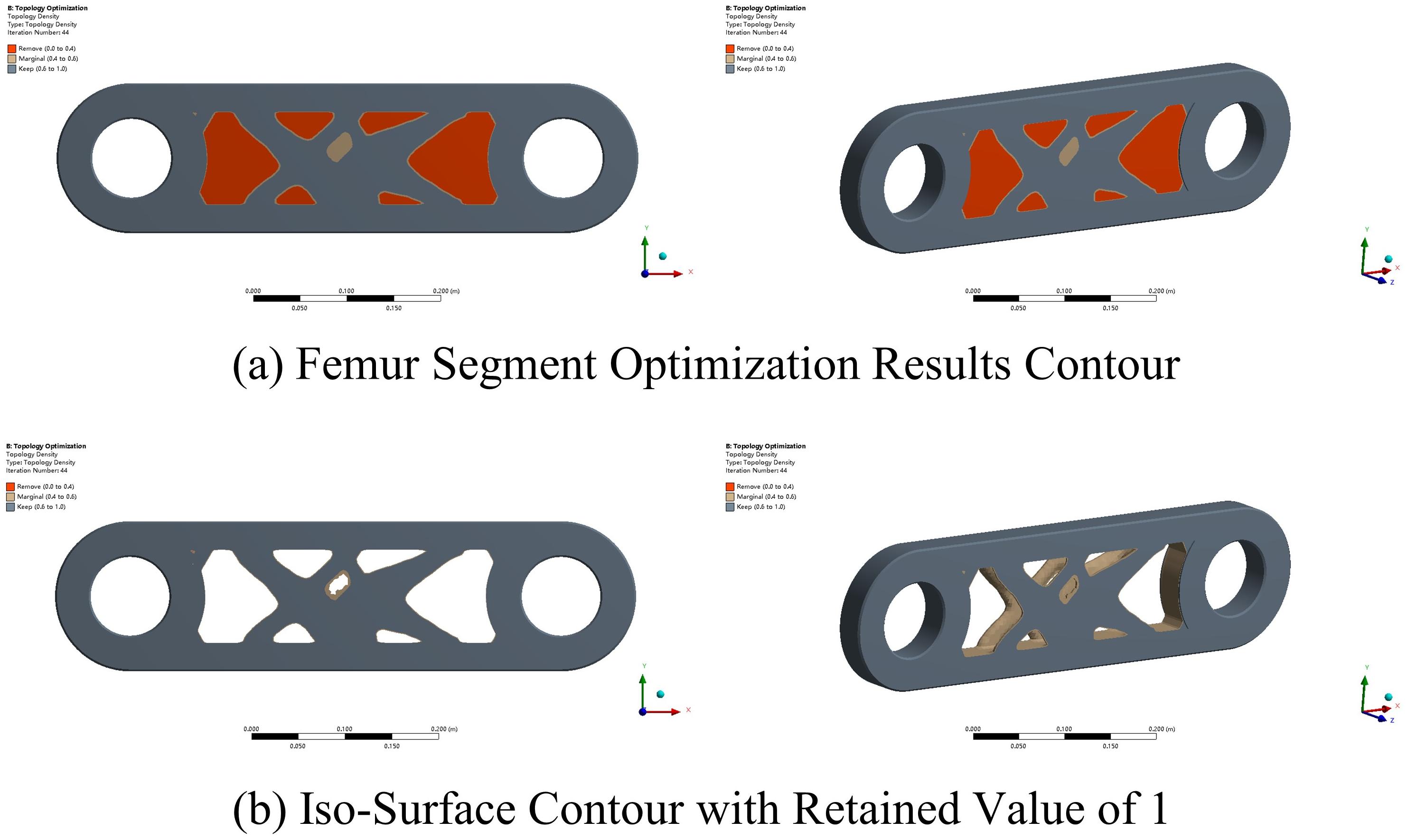}
\caption{Contour of Topology Optimization Results for the Femur Segment}\label{Figure 6}
\end{figure}

\subsection{Reconstruction of the Optimized Femur Segment Model}

The iterative results obtained from ANSYS Workbench represent theoretical calculations based on the finite element method. The cross-sections consist of numerous irregularly shaped meshes, and considering the practical challenges of manufacturing, smoothing and structural modifications are required. Based on the topology optimization results, the femur segment model is reconstructed. Considering the forward and backward movement of the construction robot, a symmetric design was adopted for the side plates of the femur during the redesign.

The revised femur structure model and the overall leg structure model are shown in Figure 7.

\begin{figure}[h]
\centering
\includegraphics[width=0.6\textwidth]{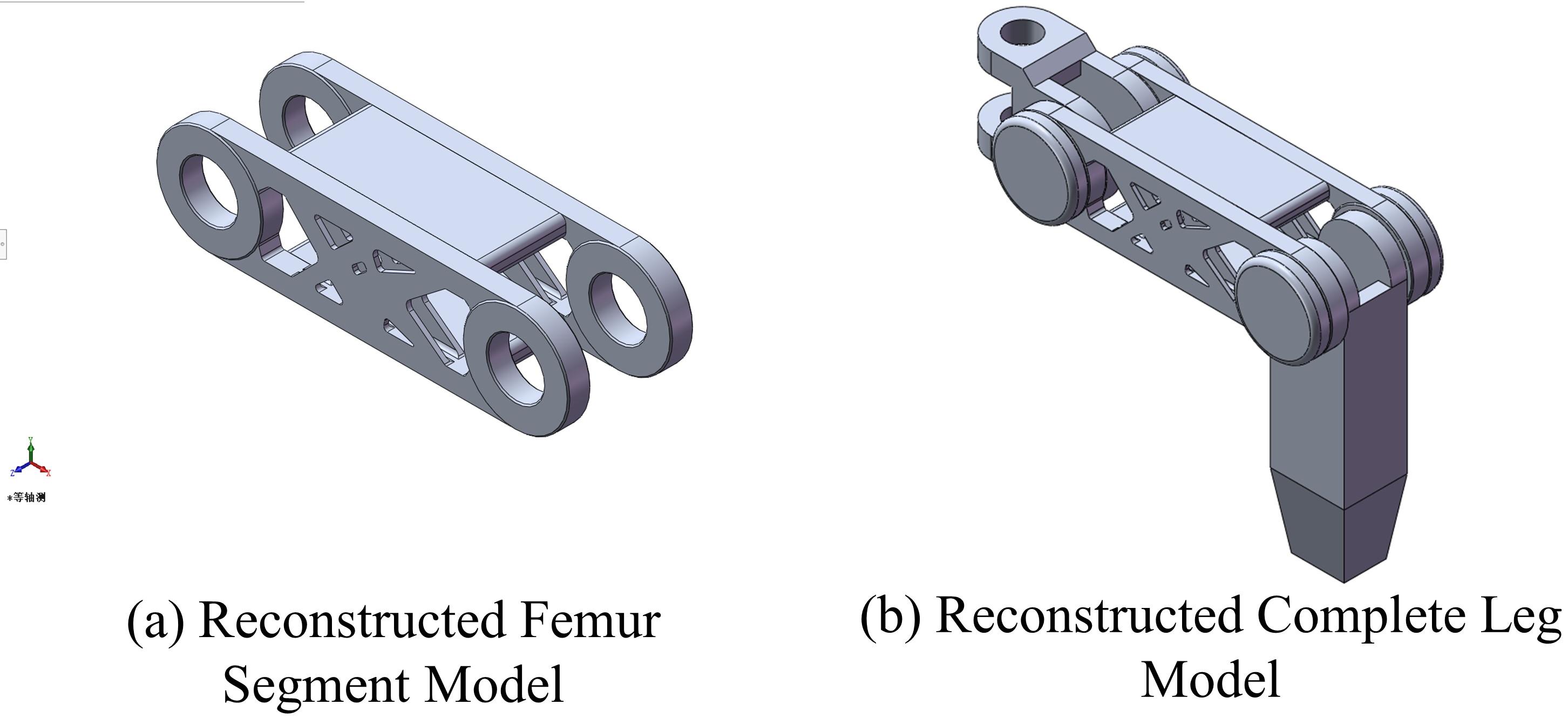}
\caption{Reconstructed Model After Optimization}\label{Figure 7}
\end{figure}

The masses of the femur segment and the overall leg before and after optimization are measured using SolidWorks software. The mass comparison before and after the optimization design is shown in Table 2. It can be observed that the masses of both the overall leg and the femur segment are significantly reduced through topology optimization.

\begin{table}[ht]
\centering
\caption{Mass Comparison Before and After Optimization}
\label{tab:mass_comparison}
\begin{tabular}{cccc}
\toprule
Structure & Mass Before Optimization (kg) & Mass After Optimization (kg) & Mass Reduction Ratio (\%) \\
\midrule
Leg Overall & 42.16 & 38.82 & 7.92 \\
Femur Segment & 17.17 & 13.83 & 19.45 \\
\bottomrule
\end{tabular}
\end{table}

\section{Finite Element Analysis of the Leg After Optimization}\label{sec2}

\subsection{Static Analysis of the Leg After Optimization}\label{subsec2}
The optimized leg, including the femur segment, is re-imported into ANSYS Workbench for static analysis under the same extreme loading conditions. The material properties, loads, and boundary conditions are set to be identical to those used in the pre-optimization static analysis. The analysis is then performed, and the overall static analysis results of the optimized leg are shown in Figure 8.
\begin{figure}[h]
\centering
\includegraphics[width=0.6\textwidth]{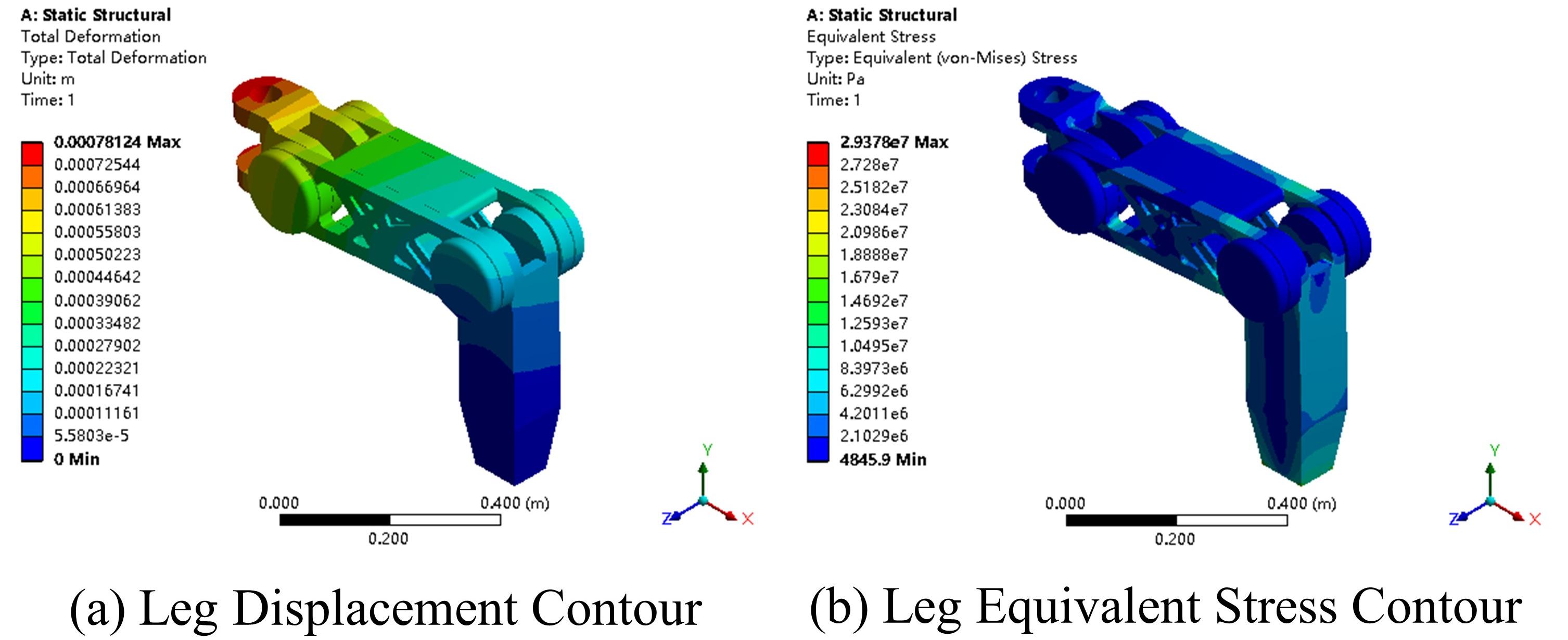}
\caption{Static Analysis Results of the Optimized Leg}\label{Figure 8}
\end{figure}

The static analysis results of the optimized femur section are shown in Figure 9.

\begin{figure}[h]
\centering
\includegraphics[width=0.6\textwidth]{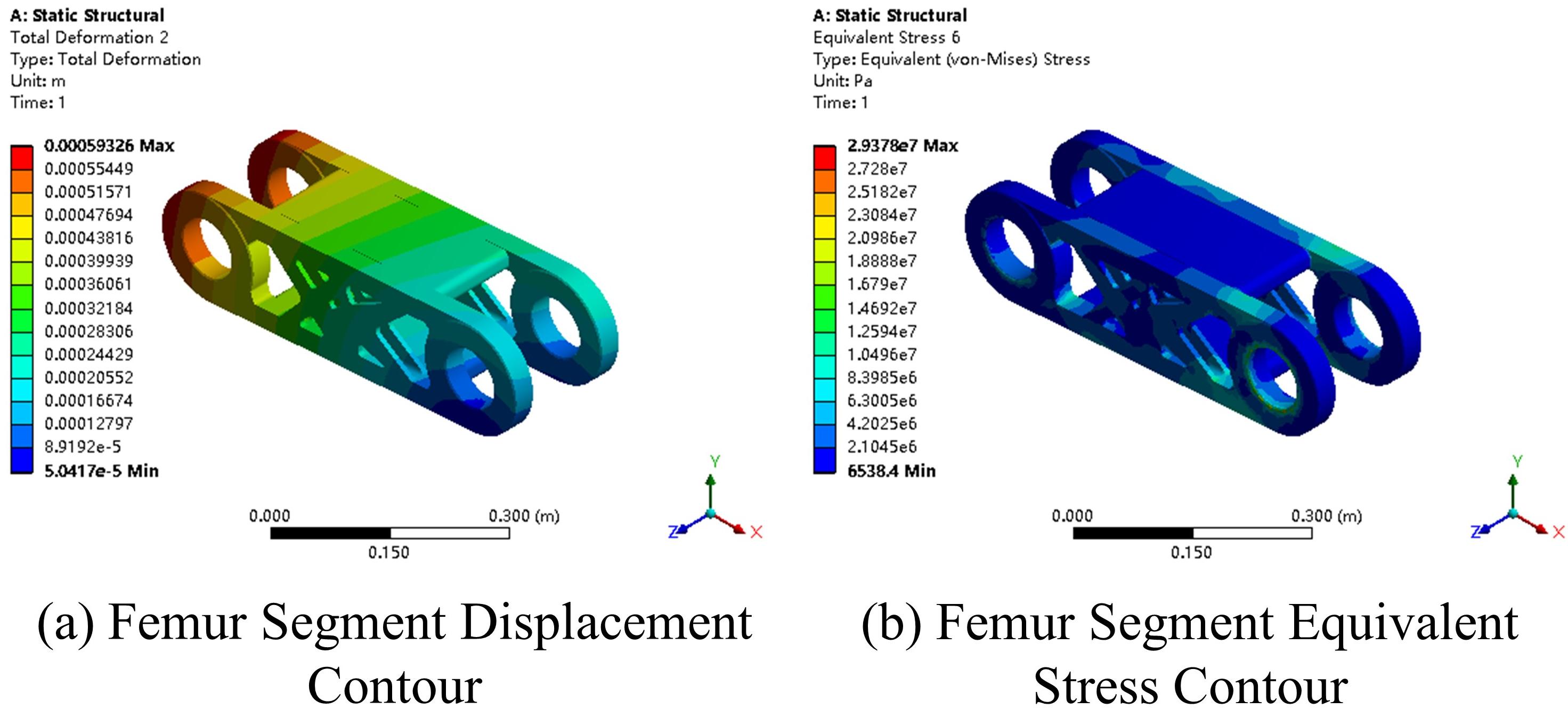}
\caption{Static Analysis Results of the Optimized Femur Segment}\label{Figure 9}
\end{figure}

The analysis results indicate that the maximum equivalent stress of the optimized leg structure is 29.38 MPa, with the femur segment experiencing the same maximum equivalent stress of 29.38 MPa, meaning the highest stress occurs in the femur. The maximum stress singularity remains at the connection points between structural components. The maximum displacement of the leg is 0.78 mm, while the maximum displacement of the femur segment is 0.59 mm. A comparison of the static performance of the overall leg and the femur segment before and after optimization is presented in Table 3.

\begin{table}[ht]
\centering
\caption{Comparison of Static Performance Before and After Optimization}
\label{tab:static_performance}
    \begin{tabular}{ccccc}
    \toprule
    \textbf{Structure} & 
    \textbf{\makecell{Stress Before \\ Optimization (MPa)}} & 
    \textbf{\makecell{Stress After \\ Optimization(MPa)}} & 
    \textbf{\makecell{Displacement Before \\ Optimization(mm)}} & 
    \textbf{\makecell{Displacement After \\ Optimization(mm)}} \\
    \midrule
    Leg Overall & 27.52 & 29.38 & 0.67 & 0.78 \\
    Femur Segment & 18.56 & 29.38 & 0.50 & 0.59 \\
    \bottomrule
    \end{tabular}%
\end{table}
It can be observed that the maximum equivalent stress and displacement of both the overall leg structure and the femur segment have increased compared to the pre-optimization design. However, they still satisfy the strength and precision requirements.

\subsection{Modal Analysis of the Optimized Leg}

Modal analysis of the optimized leg is performed under the same operating conditions, with identical material properties, loads, and boundary conditions as those used in the pre-optimization modal analysis. The first six mode shape contour plots of the optimized leg are shown in Figure 10.

\begin{figure}[h]
\centering
\includegraphics[width=0.6\textwidth]{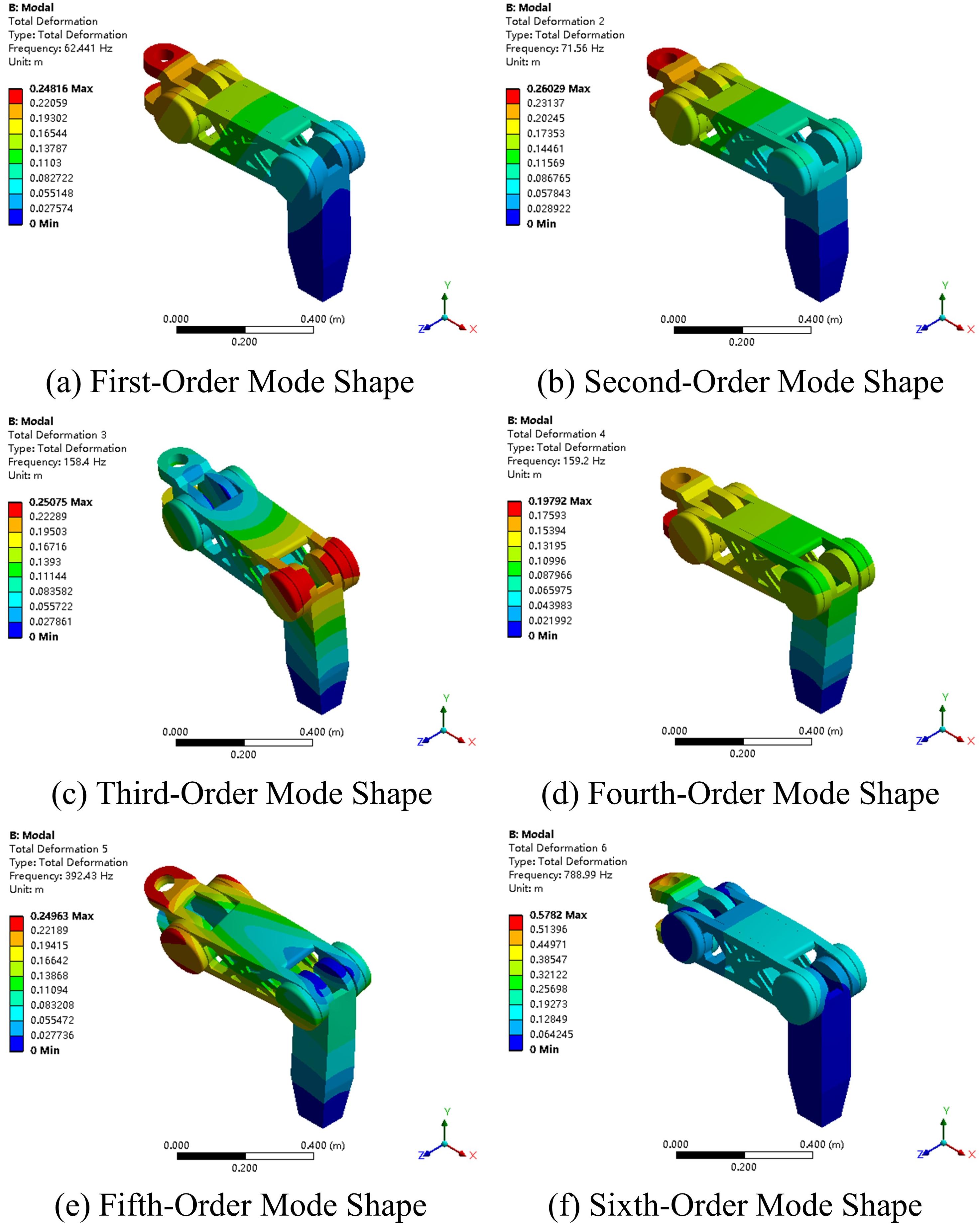}
\caption{First Six Mode Shapes of the Optimized Leg from Modal Analysis}\label{Figure 10}
\end{figure}

Based on the analysis results from the contour plots, the natural frequencies and mode shapes of the first six modes of the optimized leg are obtained and presented in Table 4.

\begin{table}[ht]
\centering
\caption{First Six Mode Shapes of the Optimized Leg}
\label{tab:mode_shapes}
\begin{tabular}{cccc}
\toprule
\textbf{Mode Number} & 
\textbf{Natural Frequency (Hz)} & 
\textbf{Vibration Type} \\
\midrule
1 & 62.441 & Bending in XOZ Plane \\
2 & 71.56 & Bending in XOY Plane \\
3 & 158.4 & Combined Bending and Torsion \\
4 & 159.2 & Bending in YOZ Plane \\
5 & 392.43 & Combined Bending and Torsion \\
6 & 788.99 & Combined Bending and Torsion \\
\bottomrule
\end{tabular}
\end{table}

Based on Table 1 and Table 4, the bar chart in Figure 11 illustrates the changes in the natural frequencies of the first six modes of the leg before and after optimization. It can be observed that the natural frequency values of each mode decrease after optimization. However, for the optimized leg, the natural frequency of the first mode is 62.441 Hz, and all subsequent modes still exhibit natural frequencies higher than the operating frequency of the leg. Therefore, resonance is unlikely to occur during the leg's operation.

\begin{figure}[h]
\centering
\includegraphics[width=0.6\textwidth]{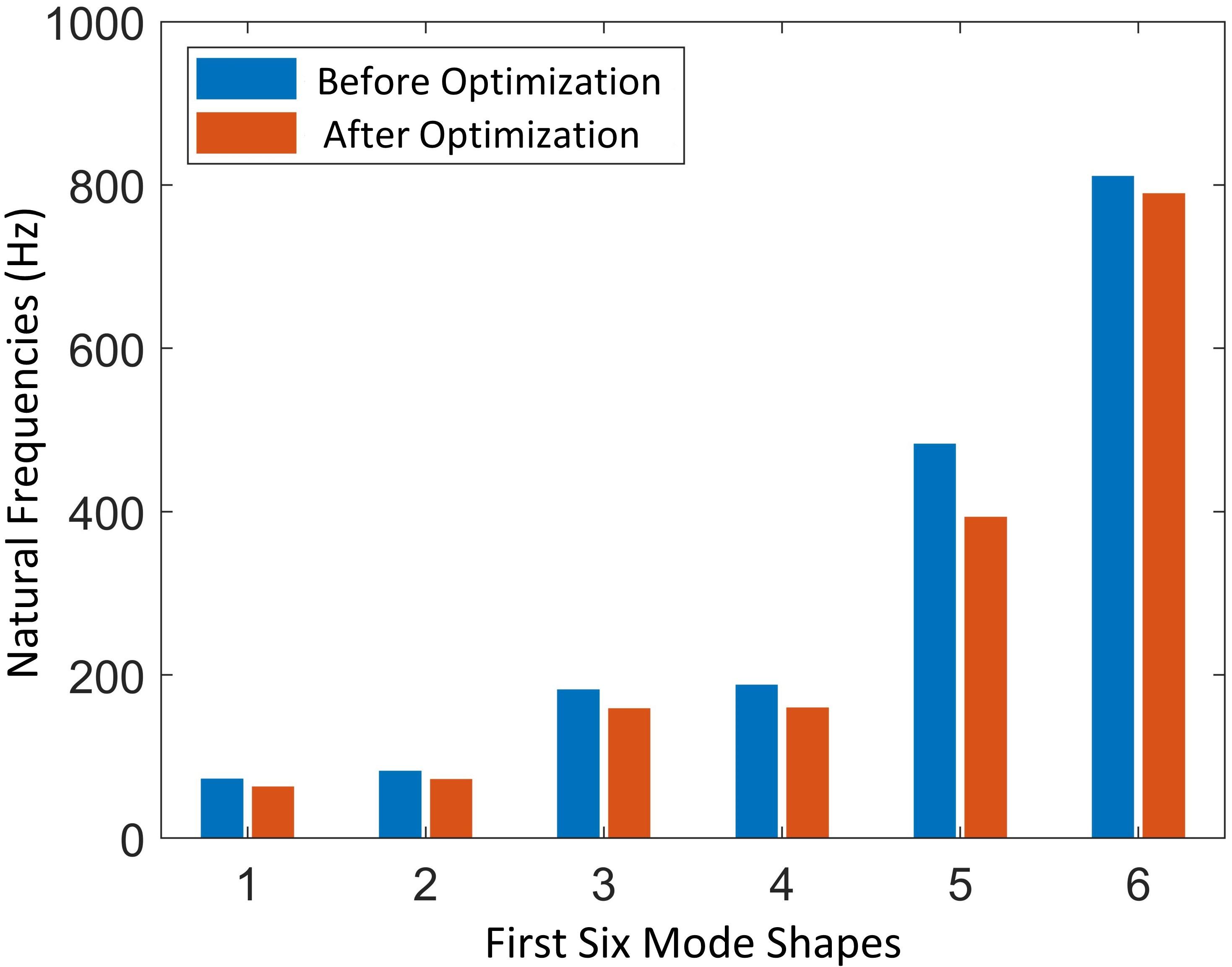}
\caption{First Six Mode Frequencies of the Leg Before and After Optimization}\label{in Figure 11}
\end{figure}

The results of the static and modal analyses show that the optimized leg structure continues to satisfy the static and dynamic performance standards, thereby validating the reasonableness of the topology optimization design presented in this paper.

\section{Conclusion}

This study addresses the lightweight requirements of construction robots in building environments by proposing a topology optimization and secondary structural reconstruction strategy for the leg based on the SIMP variable density method. The design's rationality is validated through finite element analysis using ANSYS Workbench, which performs static and modal analyses on the supporting leg. Based on the finite element analysis results, the SIMP-based variable density method is applied to conduct targeted topology optimization on the femur segment, which accounts for the highest proportion of the leg's weight. After iterative calculations and secondary structural reconstruction, a mass reduction of 19.45\% in the femur and 7.92\% in the overall leg is achieved. Furthermore, static and modal analyses performed on the reconstructed leg model demonstrate that the optimized leg maintains good structural stability and dynamic performance, confirming the effectiveness and feasibility of the proposed lightweight design approach.

Although this study has achieved significant results under safety considerations, future work could explore combining multiple topology optimization methods to achieve deeper structural optimization. This would not only provide stronger technical support for the design of construction robots but also lay the foundation for more flexible and efficient construction robots. In summary, this study not only opens new pathways for improving the payload-to-weight ratio of construction robots but also provides important theoretical and technical support for their application in complex construction environments.

\section*{Statements and Declarations}
\textbf{Conflict of Interest Statement} \\
On behalf of all authors, the corresponding author declares that there are no competing interests.

\bigskip 

\noindent \textbf{Replication of Results} \\
The study employs ANSYS Workbench for finite element static and modal analyses of a robotic leg structure. The primary material, 6061 aluminum alloy, is defined with a density of 2.70 g/cm³, Poisson’s ratio of 0.33, elastic modulus of 69 GPa, yield strength of 276 MPa, and a safety factor of 2. A global gravitational acceleration of 9.80665 m/s² is applied to simulate real-world operational conditions. All boundary conditions, material properties, and meshing parameters are described in sufficient detail in the main text to allow replication of the analysis using the ANSYS Workbench platform. However, the original ANSYS project files are not publicly available due to institutional data-sharing policies.

\bigskip 
\noindent \textbf{Funding} \\
This work was supported by the National Key R\&D Program of China under Grant No. 2023YFB4705002; the National Natural Science Foundation of China under Grant No. U20A20283; the Guangdong Provincial Key Laboratory of Construction Robotics and Intelligent Construction under Grant No. 2022KSYS013; the CAS Science and Technology Service Network Plan (STS) - Dongguan Special Project under Grant No. 20211600200062; and the Science and Technology Cooperation Project of Chinese Academy of Sciences in Hubei Province Construction 2023.

\bigskip 
\noindent \textbf{Author Contributions} \\
Xiao Liu: Conceptualization, Methodology, Writing – review \& editing. \\
Xianlong Yang: Software, Validation, Writing – original draft.\\
Weijun Wang: Supervision, Visualization, Investigation.\\
Wei Feng: Funding acquisition, Project administration.

\bigskip 
\noindent \textbf{Ethics Approval and Consent to Participate} \\
Not applicable. This research did not involve human participants or animals.

\bigskip 
\noindent \textbf{Data Availability} \\
The datasets generated during and/or analyzed during the current study are available from the corresponding author upon reasonable request.


\begin{thebibliography}{99}
\bibitem{c1}
N. Melenbrink, J. Werfel, and A. Menges, ``On-site autonomous construction robots: Towards unsupervised building,'' \textit{Automation in Construction}, 2020, 119: 103312. https://doi.org/10.1016/j.autcon.2020.103312

\bibitem{c2}
N. K. Muthumanickam, N. Brown, J. P. Duarte, et al., ``Multidisciplinary design optimization in Architecture, Engineering, and Construction: a detailed review and call for collaboration,'' \textit{Structural and Multidisciplinary Optimization}, 2023, 66(11): 239. https://doi.org/10.1007/s00158-023-03673-y

\bibitem{c3}
B. Xiao, C. Chen, and X. Yin, ``Recent advancements of robotics in construction,'' \textit{Automation in Construction}, 2022, 144: 104591. https://doi.org/10.1016/j.autcon.2022.104591

\bibitem{c4}
S. Tong, W. Xu, X. Zhang, et al., ``Experimental and theoretical analysis on truss construction robot: automatic grasping and hoisting of concrete composite floor slab,'' \textit{Journal of Field Robotics}, 2023, 40(2): 272-288.  https://doi.org/10.1002/rob.22128

\bibitem{c5}
L. Yang, B. Li, J. Feng, et al., ``Automated wall-climbing robot for concrete construction inspection,'' \textit{Journal of Field Robotics}, 2023, 40(1): 110-129.  https://doi.org/10.1002/rob.22119

\bibitem{c6}
D. Guan, N. Yang, J. Lai, et al., ``Kinematic modeling and constraint analysis for robotic excavator operations in piling construction,'' \textit{Automation in Construction}, 2021, 126: 103666. https://doi.org/10.1016/j.autcon.2021.103666

\bibitem{c7}
R. Fukui, Y. Yamada, K. Mitsudome, et al., ``Hangrawler: Large-payload and high-speed ceiling mobile robot using crawler,'' \textit{IEEE Transactions on Robotics}, 2020, 36(4): 1053-1066. https://doi.org/10.1109/TRO.2020.2973100

\bibitem{c8}
V. Ferravante, E. Riva, M. Taghavi, et al., ``Dynamic analysis of high precision construction cable-driven parallel robots,'' \textit{Mechanism and Machine Theory}, 2019, 135: 54-64. https://doi.org/10.1016/j.mechmachtheory.2019.01.023

\bibitem{c9}
K. Iturralde, M. Feucht, D. Illner, et al., ``Cable-driven parallel robot for curtain wall module installation,'' \textit{Automation in Construction}, 2022, 138: 104235. https://doi.org/10.1016/j.autcon.2022.104235

\bibitem{c10}
J. Liu, A. T. Gaynor, S. Chen, et al., ``Current and Future Trends in Topology Optimization for Additive Manufacturing,'' \textit{Structural and Multidisciplinary Optimization}, 2018, 57(6): 2457-2483. 
https://doi.org/10.1007/s00158-018-1994-3

\bibitem{c11}
I. Kuznetsov, V. Kozhin, A. Novokshenov, et al., ``Optimization approach for the core structure of a wind turbine blade,'' \textit{Structural and Multidisciplinary Optimization}, 2024, 67(7): 108. https://doi.org/10.1007/s00158-024-03812-z

\bibitem{c12}
Y. Lu and L. Tong, ``Topology optimization of compliant mechanisms and structures subjected to design-dependent pressure loadings,'' \textit{Structural and Multidisciplinary Optimization}, 2021, 63: 1889-1906. https://doi.org/10.1007/s00158-020-02786-y

\bibitem{c13}
T. Ho-Nguyen-Tan and H. G. Kim, ``An efficient method for shape and topology optimization of shell structures,'' \textit{Structural and Multidisciplinary Optimization}, 2022, 65(4): 119. https://doi.org/10.1007/s00158-022-03213-0

\bibitem{c14}
S. Sorohan and D. M. Constantinescu, ``Stress uniformization using functionally graded materials,'' \textit{The Romanian Journal of Technical Sciences. Applied Mechanics.}, 2021, 66(2): 163-177. 

\bibitem{c15}
T. Zuo, H. Han, Q. Wang, et al., ``An explicit topology and thickness control approach in SIMP-based topology optimization,'' \textit{Computers \& Structures}, 2025, 307: 107631. https://doi.org/10.1016/j.compstruc.2024.107631

\bibitem{c16}
H. L. Simonetti, V. S. Almeida, F. A. Neves, et al., ``3D Structural Topology Optimization Using ESO, SESO and SERA: Comparison and an Extension to Flexible Mechanisms,'' \textit{Applied Sciences}, 2023, 13(10): 6215. https://doi.org/10.3390/app13106215

\bibitem{c17}
N. Wei, H. Ye, X. Zhang, et al., ``Lightweight topology optimization of graded lattice structures with displacement constraints based on an independent continuous mapping method,'' \textit{Acta Mechanica Sinica}, 2022, 38(4): 421352. https://doi.org/10.1007/s10409-021-09047-x

\bibitem{c18}
H. Ye, Z. Dai, W. Wang, et al., ``Icm Method for Topology Optimization of Multimaterial Continuum Structure with Displacement Constraint,'' \textit{Acta Mechanica Sinica}, 2019, 35(3): 552-562. https://doi.org/10.1007/s10409-018-0827-3

\bibitem{c19}
F. H. Scherer, M. Zarroug, H. Naceur, et al., ``Topology optimization of curved thick shells using level set method and non-conforming multi-patch isogeometric analysis,'' \textit{Computer Methods in Applied Mechanics and Engineering}, 2024, 430: 117205. https://doi.org/10.1016/j.cma.2024.117205

\bibitem{c20}
W. Xie, Q. Xia, Q. Yu, et al., ``An effective phase field method for topology optimization without the curvature effects,'' \textit{Computers \& Mathematics with Applications}, 2023, 146: 200-212. https://doi.org/10.1016/j.camwa.2023.06.037 

\bibitem{c21}
Z. Li, H. Xu, and S. Zhang, ``A comprehensive review of explicit topology optimization based on moving morphable components (MMC) method,'' \textit{Archives of Computational Methods in Engineering}, 2024, 31(5): 2507-2536. https://doi.org/10.1007/s11831-023-10053-8

\bibitem{c22}
A. A. Novotny, C. G. Lopes, and R. B. Santos, ``Topological derivative-based topology optimization of structures subject to self-weight loading,'' \textit{Structural and Multidisciplinary Optimization}, 2021, 63: 1853-1861. https://doi.org/10.1007/s00158-020-02780-4

\bibitem{c23}
Y. Sun and T. C. Lueth, ``Enhancing torsional stiffness of continuum robots using 3-D topology optimized flexure joints,'' \textit{IEEE/ASME Transactions on Mechatronics}, 2023, 28(4): 1844-1852. https://doi.org/10.1109/TMECH.2023.3266873

\bibitem{c24}
L. Sha, A. Lin, X. Zhao, et al., ``A Topology Optimization Method of Robot Lightweight Design Based on the Finite Element Model of Assembly and Its Applications,'' \textit{Science Progress}, 2020, 103(3): 2147483647. https://doi.org/10.1177/0036850420936482

\bibitem{c25}
H. Ding, Z. Shi, Y. Hu, et al., ``Lightweight design optimization for legs of bipedal humanoid robot,'' \textit{Structural and Multidisciplinary Optimization}, 2021, 64(4): 2749-2762. https://doi.org/10.1007/s00158-021-02968-2

\bibitem{c26}
J. Wu, O. Sigmund, and J. P. Groen, ``Topology optimization of multi-scale structures: a review,'' \textit{Structural and Multidisciplinary Optimization}, 2021, 63: 1455-1480. https://doi.org/10.1007/s00158-021-02881-8

\bibitem{c27}
Y. Sun, Y. Liu, F. Pancheri, et al., ``Larg: A lightweight robotic gripper with 3-d topology optimized adaptive fingers,'' \textit{IEEE/ASME Transactions on Mechatronics}, 2022, 27(4): 2026-2034. https://doi.org/10.1109/TMECH.2022.3170800

\bibitem{c28}
G. Vantyghem, W. De Corte, M. Steeman, et al., ``Density-based topology optimization for 3D-printable building structures,'' \textit{Structural and Multidisciplinary Optimization}, 2019, 60: 2391-2403. https://doi.org/10.1007/s00158-019-02330-7

\bibitem{c29}
S. Zhang, H. Li, and Y. Huang, ``An improved multi-objective topology optimization model based on SIMP method for continuum structures including self-weight,'' \textit{Structural and Multidisciplinary Optimization}, 2021, 63: 211-230. https://doi.org/10.1007/s00158-020-02685-2

\end{thebibliography}
\end{document}